\documentclass[10pt,twocolumn,letterpaper]{article}
\IfFileExists{cvpr.sty}{\usepackage[review]{cvpr}}{\usepackage[letterpaper,margin=0.75in]{geometry}\usepackage{times}}
\usepackage{graphicx,booktabs,amsmath,amssymb,microtype,enumitem,hyperref,etoolbox}
\AtBeginEnvironment{thebibliography}{\setlength{\itemsep}{0pt}\setlength{\parskip}{0pt}}
\hypersetup{hidelinks}

\title{Multimodal Cross-Condition Learning for\\Temporal Sentence Grounding in Videos}
\author{Allen He\thanks{Work primarily done during internship at Stanford University.}\\BASIS International School Park Lane Harbour}
\date{}
\begin{document}
\maketitle

\begin{abstract}
Temporal sentence grounding requires a video representation that preserves query-relevant evidence and a sentence representation that reflects the visual context in which it is interpreted. Prior end-to-end grounding addresses the first requirement through sentence-conditioned visual adaptation, but leaves the language representation independent of the video. This work proposes Multimodal Cross-Condition Learning (MCCL). Its Video Prefix Adapter (VPA) converts pooled visual context into a sequence of continuous tokens that condition a frozen DistilBERT encoder. Combined with the Sentence Conditioned Adapter (SCADA), VPA supports an ordered exchange from vision to language and back to vision. A comprehensive study examines two video backbones, LSTM and Mamba temporal modules, and four conditioning modes on two datasets. Bidirectional conditioning exceeds the stronger one-way mode by 1.6--2.3 mIoU points in all four Charades-STA settings. ActivityNet Captions exhibits a different pattern: bidirectionality helps I3D--Mamba, whereas video-to-text conditioning is stronger with C3D. The study extends one-way visual adaptation into a configurable multimodal framework and reports descriptive gains over the corresponding baselines.
\end{abstract}

\section{Introduction}
Temporal sentence grounding in videos (TSGV), also called video moment retrieval, identifies the time interval described by a natural language query~\cite{gao2017tall,hendricks2017localizing}. Unlike temporal action detection with a fixed category vocabulary, the target is specified by a sentence that can distinguish objects, changes of state, and relations between events. A useful grounding system must therefore answer both which event matches the description and where that event begins and ends. This combination supports language-based navigation of long videos without requiring a user to inspect every frame.

Many grounding systems extract pretrained visual features offline and learn the subsequent multimodal interaction and localization modules~\cite{zhang2020tan,zhang2020vslnet,lei2021momentdetr}. This design separates expensive feature extraction from detector training, but also fixes what visual evidence the detector can access. Recognition pretraining encourages representations that remain stable across different instances of an action, while grounding additionally requires distinctions between neighboring moments within the same instance. A downstream head can select and combine available evidence, but cannot recover details that the representation no longer retains.

Previous work addresses this discrepancy by optimizing the visual grounding pathway online and introducing the Sentence Conditioned Adapter (SCADA)~\cite{he2026paradigm}, which injects sentence-derived modulation into intermediate visual features.

However, the encoded sentence representation remains independent of the input video. Consider an illustrative query such as ``the person puts the object down before leaving.'' Visual feedback could help contextualize the sentence before it guides localization.

\begin{figure}[t]
\centering
\includegraphics[width=\columnwidth]{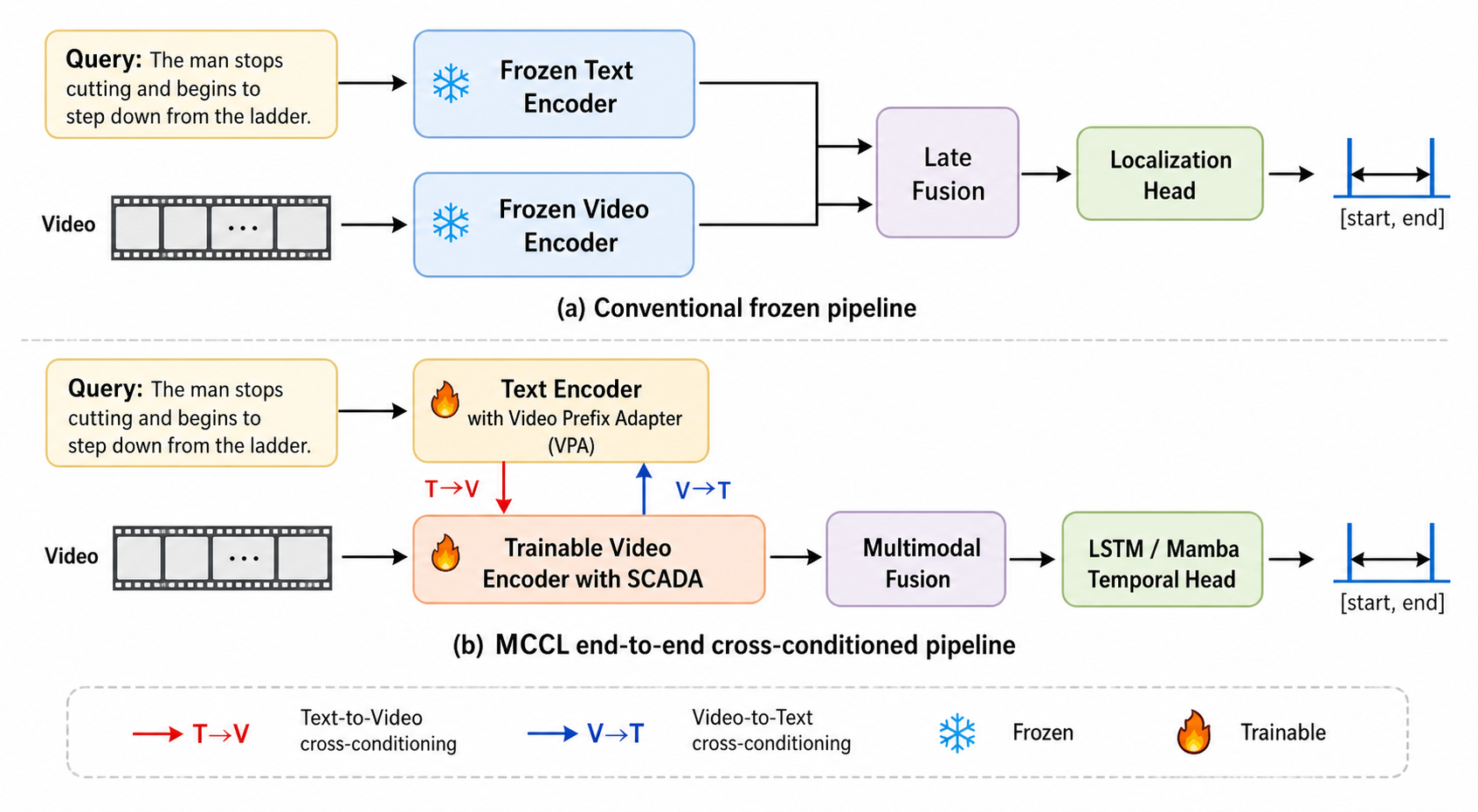}
\caption{From frozen independent encoding to end-to-end multimodal cross-conditioning. Conventional TSGV pipelines extract fixed video and text representations and interact only through late fusion. MCCL instead optimizes the grounding path end-to-end: SCADA carries text-to-video guidance, VPA carries video-to-text guidance, and an interchangeable LSTM/Mamba temporal head predicts the target boundaries.}
\label{fig:pipeline-comparison}
\end{figure}

We introduce MCCL to study this extension. A Video Prefix Adapter (VPA) maps visual context to a sequence of continuous tokens prepended to the query embeddings of DistilBERT~\cite{sanh2019distilbert}. In training, the language encoder remains frozen, while the prefix generator is trained through the localization objective. VPA supplies a video-to-text path that can operate alone or precede the text-to-video SCADA path. This yields four settings: Baseline (B), SCADA only (T), VPA only (V), and both paths (VT). Figure~\ref{fig:pipeline-comparison} places this design within the evolution from frozen features to multimodal cross-condition adaptation.

We also examine whether these conditioning choices depend on the temporal operator. LSTM and Mamba~\cite{hochreiter1997lstm,gu2023mamba} are evaluated using the same fused input and localization objective. The resulting study contains multiple dataset--backbone--temporal--conditioning configurations, plus cross-dataset evaluation. Its purpose is to examine component interactions, including failure cases. In the reported results, VT exceeds the stronger one-way path on Charades-STA, but the preferred ActivityNet direction changes with the backbone. This motivates a configurable family of conditioning paths rather than an unconditional preference for two-way feedback.

The contributions of this work are as follows:
\begin{itemize}[leftmargin=*,nosep]
\item \textbf{Video-conditioned language adaptation.} VPA generates input-level visual prefixes for a frozen language encoder, adding the video-to-text direction absent from the previous work.
\item \textbf{A unified conditioning study.} We formulate the language-vision conditioning modes within an end-to-end pipeline and examine their interaction with C3D/I3D and LSTM/Mamba. Paired effects distinguish the contribution of a second path from improvements already achieved by one-way adaptation.
\item \textbf{An analysis of domain transfer.} Cross-dataset evaluation and metric-level comparisons reveal regimes where stronger source-domain conditioning does not improve transfer, identifying concrete directions for more robust multimodal adaptation.
\end{itemize}

\section{Related Work}
\subsection{Localization and Representation Adaptation}
\paragraph{From action detection to sentence grounding.}
Temporal action detection predicts intervals together with predefined action categories. End-to-end detection studies how visual-backbone optimization and input resolution affect localization~\cite{liu2022e2etad}. This study reuses the architecture of the detector and concentrates on information exchange between video and language.

\paragraph{Temporal sentence grounding.}
Early two-stage approaches generate and rank temporal proposals~\cite{gao2017tall,hendricks2017localizing}. Later work develops local--global interaction~\cite{mun2020lgi}, dense boundary regression~\cite{zeng2020drn}, 2D temporal maps~\cite{zhang2020tan}, span prediction~\cite{zhang2020vslnet}, and graph matching~\cite{soldan2021vlg}. XML extends retrieval to long videos with aligned subtitles~\cite{lei2020xml}, while Moment-DETR removes hand-crafted proposals through set prediction~\cite{lei2021momentdetr}. Recent models strengthen query-dependent decoding and unified grounding~\cite{moon2023qddetr,lin2023univtg}. These methods principally improve interaction and decoding after feature extraction; MCCL instead adapts representations inside both modality branches.

\paragraph{Task-adaptive multimodal representations.}
Large-scale image--text pretraining demonstrates that language supervision can produce transferable visual representations~\cite{radford2021clip}, but temporal boundaries still require task-specific adaptation. Fully end-to-end TSGV and SCADA condition intermediate visual layers on the sentence~\cite{he2026paradigm}. Parameter-efficient tuning methods, including adapters~\cite{houlsby2019adapters}, LoRA~\cite{hu2022lora}, and prefix tuning~\cite{li2021prefix}, restrict which parameters are updated during adaptation. ClipCap shows that a learned visual prefix can steer a largely frozen language model for image captioning~\cite{mokady2021clipcap}; MCCL adapts this interface to temporal video evidence and boundary-sensitive grounding.

\paragraph{Prompted temporal grounding.}
Text-Visual Prompting (TVP) introduces learned textual and frame-aware visual prompts for efficient grounding with sparse 2D features~\cite{zhang2023tvp}. MCCL explores a different conditioning mechanism: VPA generates a visual prompt from the current video's representation for the sentence encoder, and SCADA modulates intermediate 3D visual features using the sentence.

\paragraph{Temporal grounding with video language models.}
TimeChat incorporates timestamp-aware frame encoding and a sliding video Q-Former for long-video understanding~\cite{ren2024timechat}. Such models broaden temporal grounding into instruction-following video tasks. MCCL instead uses a compact sentence encoder and an explicit localization head to examine representation adaptation with a fixed detector.

\subsection{Temporal Models and Generalization}
\paragraph{Long-range temporal modeling.}
LSTMs process positions recurrently~\cite{hochreiter1997lstm}; Transformers offer global interaction at quadratic attention cost. Structured state-space models provide an efficient alternative for long sequences~\cite{gu2022s4}, and Mamba introduces input-dependent selective state updates with linear-time processing~\cite{gu2023mamba}.

\paragraph{Input prompts versus layer-wise prefixes.}
Prefix tuning optimizes continuous conditioning vectors while keeping a pretrained language model fixed~\cite{li2021prefix}. Input-level prompt tuning provides a related interface by adding learned embeddings at the input~\cite{lester2021prompt}. VPA belongs to this input-level family: its prefixes are generated from the current video, rather than being a single prompt shared by every example.

\paragraph{Dataset bias and transfer.}
Temporal grounding accuracy can partly reflect annotation regularities instead of video--language reasoning. Zhang et al.~\cite{zhang2021debias} examine temporal annotation bias and study both data and model debiasing. Their findings motivate testing whether stronger in-domain conditioning also transfers across datasets. Our transfer experiments do not implement their debiasing procedures or isolate annotation bias causally. Instead, they expose a practical distinction between fitting a source benchmark and retaining useful cross-modal representations on a different target domain.

\section{Multimodal Cross-Condition Learning}
\subsection{Problem Formulation}
\begin{figure*}[t]
\centering
\includegraphics[width=\textwidth]{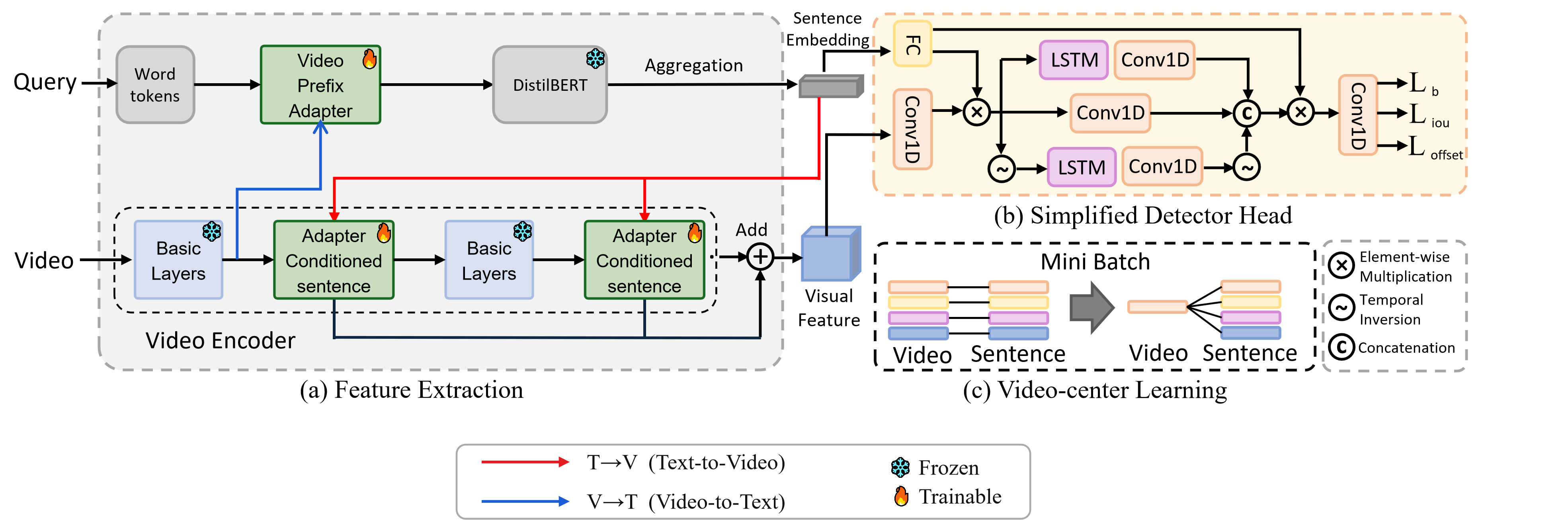}
\caption{Overall MCCL architecture. The blue video-to-text path branches from the output of the first Basic Layers, before the first sentence-conditioned adapter, and uses this query-independent visual context to generate an input-level prefix for DistilBERT. The red text-to-video path uses the resulting sentence representation to condition the visual adapters. The final aggregated visual feature is used by the temporal detector and its boundary, IoU-confidence, and offset objectives. Frozen and trainable components are marked explicitly.}
\label{fig:overview}
\end{figure*}

Given an untrimmed video $V=\{I_i\}_{i=1}^{T}$ and a query sentence $Q=\{w_j\}_{j=1}^{N}$, TSGV predicts a temporal segment $\hat{\tau}=(\hat t_s,\hat t_e)$ that maximizes its overlap with the ground-truth segment $\tau^*=(t_s^*,t_e^*)$. A video backbone samples and encodes $V$ into a temporal feature sequence $X\in\mathbb{R}^{L\times d_v}$, where $L$ is the temporal scale used by the detector. DistilBERT maps the query to token embeddings $E\in\mathbb{R}^{N\times d_e}$, with $d_e=768$, and a projected sentence representation $q\in\mathbb{R}^{d_q}$, with $d_q=512$. The model learns a scoring function
\begin{equation}
\hat{\tau}=\operatorname*{arg\,max}_{\tau\in\mathcal{T}(V)} s(\tau\mid V,Q),
\end{equation}
where $\mathcal{T}(V)$ denotes candidate temporal locations generated over the normalized timeline. Figure~\ref{fig:overview} illustrates the complete end-to-end architecture and the two cross-conditioning directions.

\subsection{End-to-End Grounding-Oriented Training}
MCCL does not rely on a separate, offline video-feature extraction stage. Frames are passed through the pretrained C3D or I3D backbone during training, which remains frozen and provides the query-independent visual context and visual features. The grounding objective updates the active cross-condition modules, temporal module, and localization heads, while the main DistilBERT weights remain frozen. Thus, end-to-end refers to the joint optimization of the trainable conditioning and localization components, rather than unrestricted fine-tuning of the pretrained backbones.

Query-independent computation from the frozen visual backbone is shared across sentence annotations associated with the same video within a mini-batch. This avoids repeatedly extracting identical visual features for different queries, while the subsequent sentence-conditioned computation remains query-specific.

\subsection{Text-to-Video Conditioning}
SCADA is from the preceding end-to-end framework~\cite{he2026paradigm}. Its inner branch modulates a residual visual pathway, while its outer branch supplies query-guided temporal evidence to the detector.

For an intermediate tensor $Z_\ell$, let $D_\ell$ and $U_\ell$ denote channel reduction and restoration, and let $g_\ell(q)$ project the sentence into the reduced channel space. Suppressing reshape operators, the inner branch is
\begin{equation}
\begin{aligned}
R_\ell & = \delta\!\left(D_\ell(Z_\ell)\right), \\
A_\ell & = \operatorname{DWConv1D}\!\left(\operatorname{Norm}\!\left(R_\ell \odot g_\ell(q)\right)\right), \\
Z_\ell^{\mathrm{inner}} & = Z_\ell + \delta\!\left(U_\ell(A_\ell)\right).
\end{aligned}
\end{equation}
The sentence projection is broadcast over visual positions. The outer branch applies analogous sentence modulation between 3D projection layers, then it spatially pools the result into temporal tokens. Compatible outer outputs are aggregated with the final backbone representation. The residual inner path preserves a direct route for pretrained visual features, whereas the outer path makes intermediate evidence available to localization.

The source of $q$ depends on the conditioning mode. In T mode, SCADA receives the ordinary sentence encoding; in VT mode, it receives the video-conditioned encoding generated through VPA.

\subsection{Video-to-Text Conditioning}
\begin{figure}[t]
\centering
\includegraphics[width=\columnwidth]{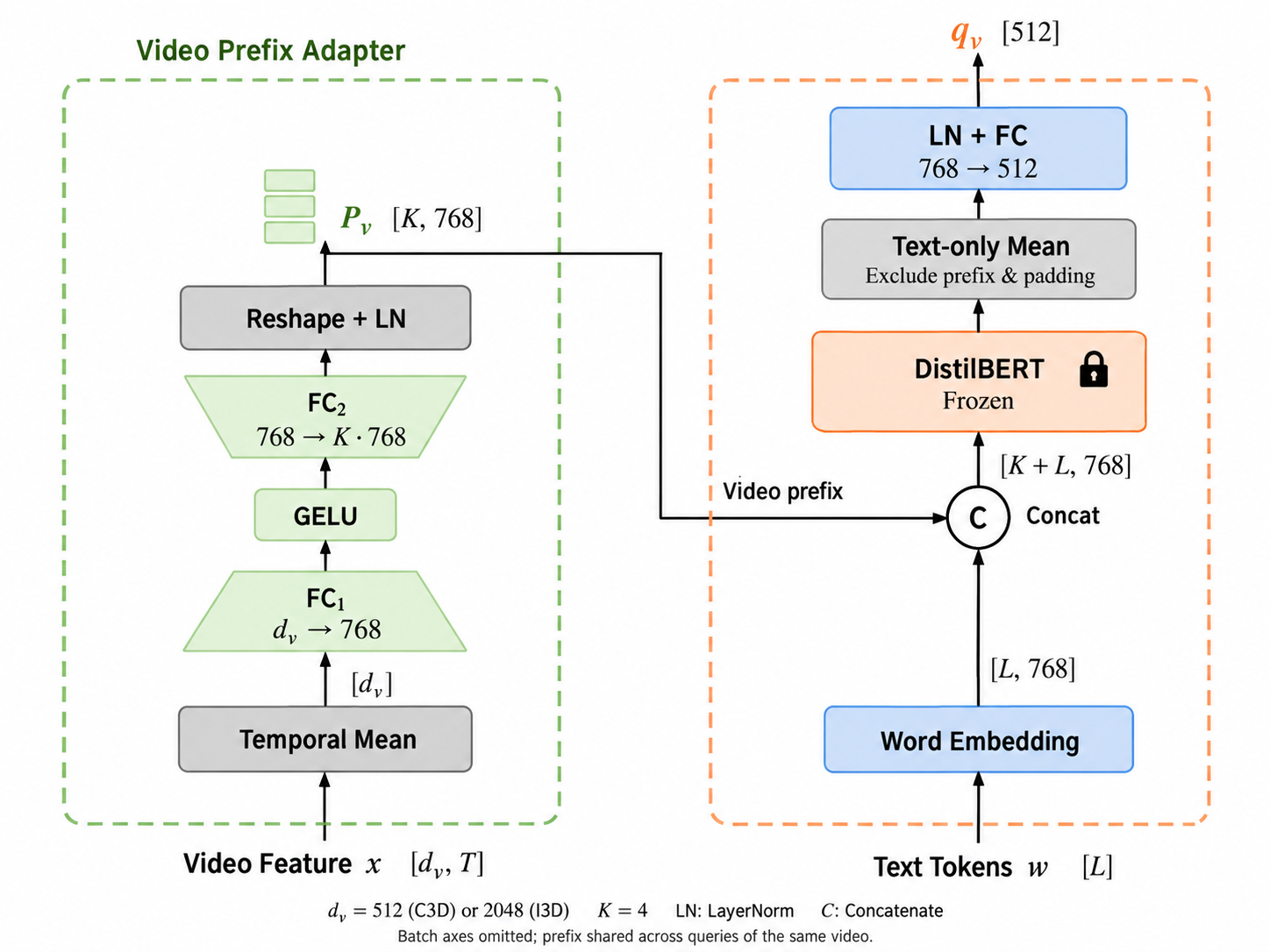}
\caption{Video Prefix Adapter and prefix injection. The C3D/I3D backbone produces a temporally aligned feature sequence with channel width $d_v\in\{512,2048\}$. Temporal mean pooling followed by the trainable FC1--GELU--FC2 projection and LayerNorm generates $K=4$ prefix tokens of width 768. The tokens are concatenated with the word embeddings and processed by frozen DistilBERT; the resulting sentence representation is projected to $q_v\in\mathbb{R}^{512}$.}
\label{fig:vpa}
\end{figure}

VPA provides the novel video-to-text direction. It is a trainable video-conditioned soft-prompt generator whose input is the query-independent visual context produced by the frozen visual backbone and whose output is a sequence of continuous prefix embeddings. Its internal projection and prefix-injection process is shown in Figure~\ref{fig:vpa}. Let $X^{\mathrm{ctx}}\in\mathbb{R}^{d_v\times T}$ denote the temporally aligned backbone feature sequence before the first sentence-conditioned adapter. For C3D, $d_v=512$; for I3D, $d_v=2048$. VPA first averages $X^{\mathrm{ctx}}$ over time and then applies a two-layer projection with hidden width 768:
\begin{equation}
P_v = \operatorname{reshape}\!\left(W_2\,\delta\!\left(W_1\operatorname{Mean}_t(X^{\mathrm{ctx}})\right)\right)\in\mathbb{R}^{K\times d_e}.
\end{equation}
With $K=4$, these tokens have shape $[K,768]$ per video and are prepended to the word embeddings before DistilBERT encoding. Let $\theta_{\mathrm{VPA}}$ denote the parameters of the FC1, FC2, reshape, and LayerNorm modules. Training updates $\theta_{\mathrm{VPA}}$ through the grounding objective. The visual tensor and the original video pixels are inputs to this generator rather than learnable parameters. Thus, VPA provides a trainable interface through which global video context influences the sentence representation.

Prefix tokens constrain video feedback to a bottleneck conditioning interface. They influence self-attention in every DistilBERT layer through the additional prefix positions, while the lexical embeddings and transformer weights remain anchored to their pretrained language distribution.

\subsection{Bidirectional Cross-Conditioning}
The $VT$ variant activates both paths in an ordered computation. VPA first produces a video-conditioned sentence representation $q_v$; SCADA then uses $q_v$ to adapt the visual features. After spatial pooling and temporal feature projection, the video and sentence features are combined by element-wise multiplication,
\begin{equation}
H = \psi_v\!\left(X_{q_v}\right) \odot \psi_q\!\left(q_v\right),
\end{equation}
where $X_{q_v}$ denotes the text-conditioned video sequence. This ordering avoids a circular computation graph while still allowing information to traverse both modalities: video evidence disambiguates the query, and the resulting query representation selects relevant visual channels and time steps.

\subsection{Information Flow and Adaptation Scope}
\paragraph{Separating visual context from the conditioned output.}
The ordered computation requires a distinction between the visual evidence available to VPA and the visual representation produced after SCADA. Denote the former by $X^{\mathrm{ctx}}$ and the latter by $X^{\mathrm{cond}}$. The dependency is
\begin{equation}
\begin{aligned}
P_v & = A_v\!\left(\operatorname{Pool}\!\left(X^{\mathrm{ctx}}\right)\right), \\
q_v & = \operatorname{Pool}_{Q}\!\left(F_q\!\left([P_v; E_Q]\right)\right), \\
X^{\mathrm{cond}} & = F_v(V; q_v).
\end{aligned}
\label{eq:ordered}
\end{equation}
Here $E_Q$ denotes lexical input embeddings, $F_q$ the frozen language encoder, and $\operatorname{Pool}_{Q}$ sentence pooling over the query representation. This notation distinguishes input embeddings from contextualized word features. VPA converts the pooled $X^{\mathrm{ctx}}$ into prefix embeddings by optimizing $\theta_{\mathrm{VPA}}$; it does not optimize the pixels or the DistilBERT weights. The final aggregated visual feature after SCADA is denoted by $X^{\mathrm{cond}}$ and is used only by the localization pathway. Consequently, the VPA path is completed before the SCADA computation that consumes $q_v$, and the ordered graph contains no circular dependency.

\paragraph{Frozen weights still transmit gradients.}
Freezing DistilBERT prevents updates to its parameters, but does not remove the differentiable path from the grounding objective to the prefix generator. If $\theta_{\mathrm{VPA}}$ denotes the VPA mapping parameters, the contribution through the language branch has the chain-rule form
\begin{equation}
\frac{\partial \mathcal{L}}{\partial \theta_{\mathrm{VPA}}}
 = \frac{\partial \mathcal{L}}{\partial q_v}\,
   \frac{\partial q_v}{\partial P_v}\,
   \frac{\partial P_v}{\partial \theta_{\mathrm{VPA}}}.
\end{equation}
In the bidirectional variant, $\partial\mathcal L/\partial q_v$ includes the direct fusion path and the indirect path through SCADA. Consequently, the VPA mapping learns which aspects of the global visual context help localization even though the language weights remain fixed.

\paragraph{A compact interface with a temporal limitation.}
If the prefix projection uses bottleneck width $r$, its two weight matrices contain $d_v r + rK d_e$ parameters, excluding biases. This expression describes the specified projection, not the total trainable model size. The $K=4$ prefixes provide a fixed-size interface regardless of video length. 

\subsection{Temporal Localization Objective}
The temporal module is implemented as either LSTM or bidirectional Mamba. LSTM carries a recurrent hidden state and provides a strong local-order baseline. Mamba processes the sequence in forward and reverse directions using selective state-space updates, after which the two outputs are concatenated and projected. We keep the fusion representation and all prediction heads unchanged, so differences can be attributed to temporal modeling rather than a different detection objective.

The detector follows the simplified boundary-aware head of the prior end-to-end framework~\cite{he2026paradigm}. Given temporal features $H\in\mathbb{R}^{L\times d}$, lightweight 1D convolutional heads predict three outputs: an IoU confidence map over candidate moments, a start/end boundary distribution, and offsets that refine coarse candidate locations. Candidate labels are assigned by temporal IoU with $\tau^*$; proposals above the positive threshold are optimized as positives, while remaining candidates provide ranking supervision. The total objective is
\begin{equation}
\mathcal{L} = \mathcal{L}_{\mathrm{cls}} + 5\,\mathcal{L}_{\mathrm{reg}} + 10\,\mathcal{L}_{\mathrm{bnd}},
\end{equation}
where $\mathcal L_{\rm cls}$ supervises candidate confidence, $\mathcal L_{\rm reg}$ refines the start/end offsets of positive candidates, and $\mathcal L_{\rm bnd}$ sharpens frame-level boundary probabilities. The positive IoU threshold is 0.9.

\section{Experiments}
\subsection{Setup}\label{sec:setup}
We evaluate on Charades-STA~\cite{gao2017tall} and ActivityNet Captions~\cite{krishna2017activitynet}, reporting Rank1@IoU0.5, Rank1@IoU0.7, Rank5@IoU0.5, and Rank5@IoU0.7, together with mIoU. We use Sports-1M-pretrained C3D~\cite{tran2015c3d} and Kinetics-400-pretrained I3D~\cite{carreira2017i3d}. Charades uses 256 frames and ActivityNet uses 384. Temporal and duration scales are 128. Models train for 10 epochs. Trainable temporal, localization, and adapter modules use a learning rate of $10^{-3}$, which is decayed by 0.1 after epoch 8. All variants use identical inference post-processing ($\sigma=0.35$). Experiments use PyTorch 2.0.1, CUDA 11.8, and eight A100 80GB GPUs.

\paragraph{Dataset context.}
The preceding study characterizes Charades-STA as a collection of short indoor videos and ActivityNet Captions as a collection of longer open-domain videos~\cite{he2026paradigm,gao2017tall,krishna2017activitynet}. It reports average video durations of 30.6 and 117.6 seconds, respectively.

C3D and I3D provide alternative visual configurations rather than fused streams. C3D produces 512-channel features through 3D convolutional processing, whereas I3D produces 2048-channel features through its inflated 3D residual backbone; both are temporally aligned to $T=128$ and mean-pooled before entering the same VPA mapping. Replacing only the temporal block evaluates LSTM versus Mamba. The comparisons use the four conditioning modes defined above and measure the effect of the implemented choices under the stated training protocol.

\paragraph{Trainable parameter scope.}
The pretrained C3D and I3D video backbones are frozen in all conditioning modes, and their parameters are not updated during training. They provide the query-independent visual context used by VPA and the visual features processed by SCADA. DistilBERT is also frozen in all modes. The VPA mapping parameters (FC1, FC2, reshape, and LayerNorm) are trainable in $V$ and $VT$, SCADA parameters are trainable in $T$ and $VT$, and the temporal module and localization heads are trainable in all modes.

\paragraph{Metric definitions and temporal precision.}
For predicted interval $\hat\tau$ and annotation $\tau^*$, temporal overlap is
\begin{equation}
\operatorname{IoU}\!\left(\hat{\tau}, \tau^*\right)
= \frac{\left|\hat{\tau}\cap\tau^*\right|}{\left|\hat{\tau}\cup\tau^*\right|}.
\end{equation}
Rank1@IoU$\alpha$ is the percentage of queries whose highest-ranked prediction satisfies $\operatorname{IoU}\geq\alpha$. Rank5@IoU$\alpha$ is the percentage of queries for which at least one of the five highest-ranked predictions satisfies the same threshold. mIoU averages the highest-ranked prediction's overlap over queries. These metrics answer related but different questions: thresholded ranking success measures whether a prediction meets a chosen tolerance, while mIoU retains changes on either side of that threshold.

For the in-domain results in Table~\ref{tab:full}, \textit{Baseline} denotes the non-end-to-end reference: video and text features are extracted offline and remain fixed while the downstream localization model is trained. The LSTM Baseline and T values are reported from the preceding SCADA study~\cite{he2026paradigm}; the Mamba Baseline values are our reimplementation of the same offline-feature protocol. T, V, and VT retain their stated end-to-end training protocol, so their differences from Baseline describe performance relative to the offline reference rather than the isolated effect of adding a conditioning module. In contrast, comparisons among T, V, and VT remain matched end-to-end comparisons. Replacing C3D with I3D changes both the backbone architecture and its pretraining, so the resulting differences cannot be attributed to either factor alone.

\subsection{Results}
\begin{table*}[!t]
\centering
\caption{In-domain results. \textit{Baseline} is the non-end-to-end reference with offline frozen video and text features. LSTM Baseline and T values are reported from~\cite{he2026paradigm}; Mamba Baseline values are our reimplementation of the same offline-feature protocol. T, V, and VT denote text-to-video, video-to-text, and bidirectional conditioning in the end-to-end MCCL pipeline; ``--'' denotes metrics that are unavailable. Values are percentages rounded to one decimal.}
\label{tab:full}
\resizebox{\textwidth}{!}{\begin{tabular}{llc|rrrrr|rrrrr}
\toprule
&&&\multicolumn{5}{c|}{Charades-STA}&\multicolumn{5}{c}{ActivityNet Captions}\\
Backbone&Temporal&Mode&\shortstack{Rank1@\\IoU0.5}&\shortstack{Rank1@\\IoU0.7}&\shortstack{Rank5@\\IoU0.5}&\shortstack{Rank5@\\IoU0.7}&mIoU&\shortstack{Rank1@\\IoU0.5}&\shortstack{Rank1@\\IoU0.7}&\shortstack{Rank5@\\IoU0.5}&\shortstack{Rank5@\\IoU0.7}&mIoU\\
\midrule
C3D&LSTM&Baseline~\cite{he2026paradigm}&--&22.0&--&33.0&32.3&--&25.3&--&57.2&42.0\\
&&T~\cite{he2026paradigm}&50.4&31.2&89.3&61.3&45.5&52.7&34.7&82.6&66.7&46.8\\
&&V&47.4&28.4&88.9&57.0&43.7&58.6&39.9&85.9&69.9&\textbf{52.0}\\
&&VT&52.5&31.3&89.9&59.1&47.1&52.9&34.4&82.2&62.2&47.0\\
\cmidrule(lr){2-13}
&Mamba&Baseline&32.2&22.9&61.6&31.4&33.3&45.9&26.0&70.8&57.7&42.1\\
&&T&51.6&32.1&89.4&59.8&46.9&53.6&35.4&82.7&67.3&47.3\\
&&V&46.9&28.6&85.9&53.8&43.2&55.9&38.6&81.9&63.7&49.5\\
&&VT&55.0&35.5&87.9&58.2&49.2&52.7&36.2&83.2&67.4&47.1\\
\midrule
I3D&LSTM&Baseline~\cite{he2026paradigm}&--&30.0&--&48.8&43.8&--&26.6&--&58.7&43.2\\
&&T~\cite{he2026paradigm}&64.6&43.7&92.7&72.3&55.9&48.5&31.4&81.1&66.3&45.7\\
&&V&64.3&44.2&91.8&75.3&56.1&47.0&29.9&77.0&61.9&45.0\\
&&VT&67.2&47.0&91.8&74.9&\textbf{57.8}&47.5&30.0&76.9&62.7&45.0\\
\cmidrule(lr){2-13}
&Mamba&Baseline&45.6&29.9&67.5&47.3&43.8&46.1&27.0&77.3&61.6&43.2\\
&&T&64.5&44.3&91.2&71.8&55.9&45.1&26.5&79.1&63.7&43.4\\
&&V&63.1&43.9&91.9&73.2&55.1&48.2&30.2&80.5&64.8&45.1\\
&&VT&66.2&46.3&92.2&73.2&57.7&48.8&33.0&81.7&69.5&46.3\\
\bottomrule
\end{tabular}}
\end{table*}

\paragraph{Overall in-domain results.}
On Charades-STA, $VT$ yields the highest mIoU for every backbone--temporal pair, showing complementarity within the matched end-to-end configurations. The strongest configuration, I3D--LSTM--$VT$, reaches 57.8 mIoU, 14.0 points above its Baseline reference (43.8 mIoU). I3D--Mamba--$VT$ is close at 57.7 mIoU, indicating that the result is not tied to a single temporal operator. With C3D, Mamba--$VT$ reaches 49.2 mIoU and exceeds LSTM--$VT$ by 2.1 points, indicating that the temporal-module effect depends on the visual configuration.

ActivityNet exhibits a more heterogeneous pattern. C3D--LSTM--$V$ is best overall at 52.0 mIoU. With I3D, Mamba--$VT$ reaches 46.3 mIoU, whereas LSTM is strongest with $T$ at 45.7. For C3D--Mamba, $VT$ reaches 47.1 mIoU, compared with 47.3 for $T$ and 49.5 for $V$, the strongest one-way setting. Thus, $VT$ is 2.4 points below the best one-way mode.

\subsection{Temporal Modeling and Optimization}
\paragraph{Conditioning changes the temporal-model comparison.}
Under Charades I3D--VT, LSTM exceeds Mamba by only 0.1 mIoU. With C3D--VT, Mamba exceeds LSTM by 2.1 points. On ActivityNet, C3D--Mamba--VT is close to its LSTM counterpart, exceeding it by 0.1 points, while I3D--Mamba--VT exceeds its LSTM counterpart by 1.3. Thus, averaging over conditioning modes would conceal the dependence on the surrounding model. These are descriptive contrasts on the recorded evaluation, not evidence that either temporal architecture is universally preferable.

\paragraph{Optimization considerations.}
The training losses decrease for all reported configurations, including ActivityNet C3D--Mamba--VT.

\subsection{Paired Effects and Complementarity}
\paragraph{A comparison within each model configuration.}
For a fixed dataset, backbone, and temporal block, let $m_B,m_T,m_V,m_{VT}$ denote the mIoU scores for the Baseline, T, V, and VT configurations in Table~\ref{tab:full}. We summarize each end-to-end configuration's difference from the offline-feature reference by
\begin{equation}
\Delta_a = m_a - m_B, \qquad a\in\{T, V, VT\},
\end{equation}
and define the incremental benefit over the stronger one-way path as
\begin{equation}
C = m_{VT} - \max\!\left(m_T, m_V\right).
\end{equation}
Positive $C$ means that activating both directions improves upon either direction alone in that matched setting.

\begin{table*}[t]
\centering
\caption{Descriptive mIoU contrasts for the in-domain results, in percentage points. $\Delta$ is relative to the non-end-to-end Baseline; $C$ measures VT minus the better of T and V among matched end-to-end configurations.}
\begin{tabular}{llrrrr}
\toprule
Dataset&Backbone--temporal&$\Delta_T$&$\Delta_V$&$\Delta_{VT}$&$C$\\
\midrule
Charades&C3D--LSTM&+13.2&+11.4&+14.8&+1.6\\
Charades&C3D--Mamba&+13.6&+9.9&+15.9&+2.3\\
Charades&I3D--LSTM&+12.1&+12.3&+14.0&+1.7\\
Charades&I3D--Mamba&+12.1&+11.3&+13.9&+1.8\\
ActivityNet&C3D--LSTM&+4.8&+10.0&+5.0&-5.0\\
ActivityNet&C3D--Mamba&+5.2&+7.4&+5.0&-2.4\\
ActivityNet&I3D--LSTM&+2.5&+1.8&+1.8&-0.7\\
ActivityNet&I3D--Mamba&+0.2&+1.9&+3.1&+1.2\\
\bottomrule
\end{tabular}
\end{table*}

\paragraph{Complementarity on Charades-STA.}
All four Charades settings have positive $C$, ranging from 1.6 to 2.3 points. The incremental effect has the same sign across both backbones and temporal modules in the reported evaluation. For I3D--LSTM, V is 12.3 points above Baseline, and adding T raises mIoU by a further 1.7 points. The result supports the usefulness of the second path without attributing the entire 14.0-point difference from Baseline to that path.

\paragraph{A backbone-dependent result on ActivityNet.}
Only I3D--Mamba retains positive complementarity on ActivityNet, with $C=1.2$. The remaining pairs have $C=-5.0$, $-2.4$, and $-0.7$. For C3D--Mamba, VT is 5.0 points above Baseline but 2.4 points below the stronger V setting. Consequently, averaging all VT results would hide both a useful I3D--Mamba regime and cases in which bidirectional conditioning trails the stronger one-way path.

\paragraph{Complementarity versus the offline reference.}
Because Baseline uses a different, non-end-to-end feature-extraction protocol, it is not a factorial no-conditioning cell for an additive interaction contrast. The $\Delta$ values therefore describe differences from the offline reference, rather than isolated main effects of T or V. We use $C$, which compares only the matched end-to-end T, V, and VT settings, for the practical question of whether the second path helps.

\paragraph{Strict overlap reveals more than the average.}
On ActivityNet with C3D--LSTM, T improves Rank1@IoU0.7 by 9.4 points relative to Baseline, whereas V improves it by 14.6 points. Thus, the two modes show different behavior at the strict overlap threshold reported in Table~\ref{tab:full}.

\subsection{Cross-Dataset Generalization}
We cross-test all 32 trained configurations without target-domain adaptation. Tables~\ref{tab:c2a} and~\ref{tab:a2c} report the complete transfer results. Because the source and target datasets differ substantially in video duration, scene composition, and temporal granularity, these results reflect a cross-domain stress test of representation transfer.

\begin{table*}[t]
\centering
\caption{All configurations for Charades-STA $\rightarrow$ ActivityNet Captions transfer (\%).}
\label{tab:c2a}
\scriptsize
\resizebox{0.88\textwidth}{!}{\begin{tabular}{llc|rrrrr}
\toprule
Backbone&Temporal&Mode&\shortstack{Rank1@\\IoU0.5}&\shortstack{Rank1@\\IoU0.7}&\shortstack{Rank5@\\IoU0.5}&\shortstack{Rank5@\\IoU0.7}&mIoU\\
\midrule
C3D&LSTM&B&24.7&13.3&\textbf{64.4}&\textbf{36.6}&\textbf{30.8}\\
I3D&LSTM&B&23.8&12.6&60.8&34.4&29.5\\
C3D&LSTM&T&24.0&12.6&62.8&36.5&30.0\\
I3D&LSTM&T&24.3&12.6&61.2&34.6&29.7\\
C3D&LSTM&V&24.6&\textbf{13.9}&62.2&33.4&29.8\\
I3D&LSTM&V&23.9&12.7&58.3&32.4&28.6\\
C3D&LSTM&VT&25.1&13.3&60.7&32.6&30.0\\
I3D&LSTM&VT&22.7&12.0&58.0&32.3&28.2\\
C3D&Mamba&B&25.3&13.3&59.7&32.7&30.6\\
I3D&Mamba&B&23.4&11.4&59.2&31.0&28.1\\
C3D&Mamba&T&\textbf{25.3}&13.1&61.1&33.6&30.5\\
I3D&Mamba&T&23.6&11.8&59.9&32.0&28.7\\
C3D&Mamba&V&24.9&13.5&61.5&33.7&29.7\\
I3D&Mamba&V&23.0&11.1&58.2&31.0&28.1\\
C3D&Mamba&VT&24.6&13.0&60.3&32.7&29.5\\
I3D&Mamba&VT&23.9&11.9&60.2&32.4&28.9\\
\bottomrule
\end{tabular}}
\end{table*}

\begin{table*}[t]
\centering
\caption{All configurations for ActivityNet Captions $\rightarrow$ Charades-STA transfer (\%).}
\label{tab:a2c}
\scriptsize
\resizebox{0.88\textwidth}{!}{\begin{tabular}{llc|rrrrr}
\toprule
Backbone&Temporal&Mode&\shortstack{Rank1@\\IoU0.5}&\shortstack{Rank1@\\IoU0.7}&\shortstack{Rank5@\\IoU0.5}&\shortstack{Rank5@\\IoU0.7}&mIoU\\
\midrule
C3D&LSTM&B&15.0&6.1&59.4&33.5&27.2\\
I3D&LSTM&B&22.3&\textbf{11.5}&52.2&29.4&25.6\\
C3D&LSTM&T&13.1&5.2&\textbf{60.7}&\textbf{34.9}&29.0\\
I3D&LSTM&T&20.5&11.1&51.4&28.4&23.5\\
C3D&LSTM&V&16.0&7.2&59.3&34.2&25.2\\
I3D&LSTM&V&17.5&8.2&47.1&22.3&20.9\\
C3D&LSTM&VT&15.9&6.6&56.4&33.0&\textbf{29.9}\\
I3D&LSTM&VT&19.6&9.8&51.7&28.2&22.7\\
C3D&Mamba&B&12.3&4.8&52.4&28.2&23.7\\
I3D&Mamba&B&\textbf{23.2}&10.7&54.8&29.0&27.1\\
C3D&Mamba&T&12.4&5.5&46.9&25.1&19.4\\
I3D&Mamba&T&19.0&8.5&46.4&26.5&23.3\\
C3D&Mamba&V&15.0&6.4&57.3&32.9&25.1\\
I3D&Mamba&V&18.6&8.8&45.9&23.3&21.9\\
C3D&Mamba&VT&6.7&2.6&29.3&13.2&10.8\\
I3D&Mamba&VT&19.5&9.7&42.5&25.2&22.0\\
\bottomrule
\end{tabular}}
\end{table*}

\paragraph{Charades-STA $\rightarrow$ ActivityNet Captions.}
The best mIoU is obtained by C3D--LSTM--B (30.8), followed closely by C3D--Mamba--B (30.6) and C3D--Mamba--T (30.5). Although C3D--Mamba--T gives the best Rank1@IoU0.5 (25.3), the C3D--LSTM--B model has the best Rank5@IoU0.5 and mIoU. This indicates that conditioning learned on short Charades clips does not always transfer to longer ActivityNet videos. C3D variants have higher mIoU than their matched I3D variants in this direction; this comparison does not hold for every recall threshold, and the cause of the transfer difference remains a hypothesis.

\paragraph{ActivityNet Captions $\rightarrow$ Charades-STA.}
The best mIoU is C3D--LSTM--VT (29.9), while the best Rank1@IoU0.5 and Rank1@IoU0.7 scores are achieved by I3D--Mamba--B (23.2) and I3D--LSTM--B (11.5), respectively. This split shows that transfer quality depends on the evaluation metric: C3D--LSTM--VT achieves the highest average overlap, whereas I3D B models achieve the best top-1 recall entries at IoU0.5 and IoU0.7. The mIoU alone does not identify which localization errors improved. The most severe failure remains C3D--Mamba--VT, which drops to 6.7 Rank1@IoU0.5 and 10.8 mIoU.

\paragraph{Cross-domain implications.}
The recorded cross-domain patterns motivate three possible explanations. First, bidirectional conditioning can learn source-specific video--language correlations and overfit the training domain. Second, Charades contains short, fine-grained indoor actions whereas ActivityNet contains longer, coarser open-domain events, creating a temporal-granularity mismatch. Third, start/end and duration priors shift across datasets. The asymmetry between the two transfer directions indicates a heterogeneous pattern in the records: different feature and temporal combinations respond differently to the target domain. More conditioning is therefore not automatically more transferable. Domain-adversarial alignment~\cite{ganin2016dann}, adapter regularization, or target-agnostic duration normalization could be investigated to retain the in-domain gains without encoding dataset-specific correlations.

\subsection{Interpreting Transfer Without Target Adaptation}
\paragraph{Separate the target benchmark from source training.}
A useful transfer comparison fixes the target dataset and architecture, then compares target-trained and source-trained checkpoints. For I3D--LSTM--VT evaluated on Charades, target training gives 57.8 mIoU, whereas ActivityNet training gives 22.7, a 35.1-point gap. For C3D--LSTM--V evaluated on ActivityNet, the corresponding scores are 52.0 and 29.8, a 22.2-point gap. These are descriptive contrasts with the named target dataset held fixed.

\paragraph{Conditioning can help fitting more than transfer.}
For Charades-trained I3D--LSTM, VT is 14.0 mIoU points above the Baseline in-domain reference, but its ActivityNet mIoU is 28.2, compared with 29.5 for the cross-domain B checkpoint reported above. This contrast is consistent with source-specific adaptation, but the two references use different training protocols.

\paragraph{A focused diagnostic sequence.}
The current results suggest three follow-up measurements. First, duration-stratified evaluation could determine whether errors concentrate in long videos or short target moments within long videos. Second, video-prefix shuffling at evaluation could test whether correct video--query correspondence matters beyond a generic prompt effect. Third, comparing boundary errors and top-ranked proposal overlap could separate temporal imprecision from event-selection failures.

\paragraph{Model selection under domain shift.}
The best entry in a target-domain table is descriptive and should not become a target-informed model-selection rule. A deployment-oriented transfer protocol would select checkpoints and conditioning modes using source validation data, then evaluate the chosen configuration once on the target. Our complete tables expose the sensitivity of all tested modes and show that the preferred direction changes with backbone and dataset.

\section{Conclusion}
This study proposed a framework for multimodal cross-adaptation across vision and language. In particular, VPA transforms visual context into a set of continuous tokens prepended to the text embeddings, while the prefix generator is trained through the localization objective. On Charades-STA, bidirectional conditioning has the highest mIoU in all four matched end-to-end backbone--temporal settings and exceeds the stronger one-way mode by 1.6--2.3 points. ActivityNet and cross-domain evaluation show that the preferred conditioning direction depends on the backbone, temporal module, and dataset. These results support studying the compatibility of conditioning paths rather than treating bidirectionality as universally beneficial.

\clearpage
{\small\bibliographystyle{plain}\bibliography{references}}

@inproceedings{gao2017tall, title={TALL: Temporal Activity Localization via Language Query}, author={Gao, Jiyang and Sun, Chen and Yang, Zhenheng and Nevatia, Ram}, booktitle={ICCV}, year={2017}}

@inproceedings{hendricks2017localizing, title={Localizing Moments in Video with Natural Language}, author={Hendricks, Lisa Anne and Wang, Oliver and Shechtman, Eli and Sivic, Josef and Darrell, Trevor and Russell, Bryan}, booktitle={ICCV}, year={2017}}

@inproceedings{zhang2020tan, title={Learning 2D Temporal Adjacent Networks for Moment Localization with Natural Language}, author={Zhang, Songyang and Peng, Houwen and Fu, Jianlong and Luo, Jiebo}, booktitle={AAAI}, year={2020}}

@inproceedings{zhang2020vslnet, title={Span-based Localizing Network for Natural Language Video Localization}, author={Zhang, Hao and Sun, Aixin and Jing, Wei and Zhou, Joey Tianyi}, booktitle={ACL}, year={2020}}

@inproceedings{lei2021momentdetr, title={Detecting Moments and Highlights in Videos via Natural Language Queries}, author={Lei, Jie and Berg, Tamara L. and Bansal, Mohit}, booktitle={NeurIPS}, year={2021}}

@inproceedings{moon2023qddetr, title={Query-Dependent Video Representation for Moment Retrieval and Highlight Detection}, author={Moon, WonJun and Hyun, Sangeek and Park, SangUk and Park, Dongchan and Heo, Jae-Pil}, booktitle={CVPR}, year={2023}}

@inproceedings{lin2023univtg, title={UniVTG: Towards Unified Video-Language Temporal Grounding}, author={Lin, Kevin Qinghong and Zhang, Pengchuan and Chen, Joya and Pramanick, Shraman and Gao, Difei and Wang, Alex Jinpeng and Yan, Rui and Shou, Mike Zheng}, booktitle={ICCV}, year={2023}}

@inproceedings{krishna2017activitynet, title={Dense-Captioning Events in Videos}, author={Krishna, Ranjay and Hata, Kenji and Ren, Frederic and Fei-Fei, Li and Niebles, Juan Carlos}, booktitle={ICCV}, year={2017}}

@inproceedings{tran2015c3d, title={Learning Spatiotemporal Features with 3D Convolutional Networks}, author={Tran, Du and Bourdev, Lubomir and Fergus, Rob and Torresani, Lorenzo and Paluri, Manohar}, booktitle={ICCV}, year={2015}}

@inproceedings{carreira2017i3d, title={Quo Vadis, Action Recognition? A New Model and the Kinetics Dataset}, author={Carreira, Joao and Zisserman, Andrew}, booktitle={CVPR}, year={2017}}

@inproceedings{sanh2019distilbert, title={DistilBERT, a Distilled Version of BERT: Smaller, Faster, Cheaper and Lighter}, author={Sanh, Victor and Debut, Lysandre and Chaumond, Julien and Wolf, Thomas}, booktitle={NeurIPS EMC2 Workshop}, year={2019}}

@article{hochreiter1997lstm, title={Long Short-Term Memory}, author={Hochreiter, Sepp and Schmidhuber, Juergen}, journal={Neural Computation}, volume={9}, number={8}, pages={1735--1780}, year={1997}}

@article{gu2023mamba, title={Mamba: Linear-Time Sequence Modeling with Selective State Spaces}, author={Gu, Albert and Dao, Tri}, journal={arXiv preprint arXiv:2312.00752}, year={2023}}

@inproceedings{houlsby2019adapters, title={Parameter-Efficient Transfer Learning for NLP}, author={Houlsby, Neil and Giurgiu, Andrei and Jastrzebski, Stanislaw and Morrone, Bruna and de Laroussilhe, Quentin and Gesmundo, Andrea and Attariyan, Mona and Gelly, Sylvain}, booktitle={ICML}, year={2019}}

@inproceedings{hu2022lora, title={LoRA: Low-Rank Adaptation of Large Language Models}, author={Hu, Edward J. and Shen, Yelong and Wallis, Phillip and Allen-Zhu, Zeyuan and Li, Yuanzhi and Wang, Shean and Wang, Lu and Chen, Weizhu}, booktitle={ICLR}, year={2022}}

@inproceedings{li2021prefix, title={Prefix-Tuning: Optimizing Continuous Prompts for Generation}, author={Li, Xiang Lisa and Liang, Percy}, booktitle={ACL-IJCNLP}, year={2021}}

@inproceedings{he2026paradigm,
  title={A Paradigm Shift: Fully End-to-End Training for Temporal Sentence Grounding in Videos},
  author={He, Allen and Liu, Qi and Liu, Kun and Liu, Xinchen and Liu, Wu},
  booktitle={Proceedings of the IEEE/CVF Conference on Computer Vision and Pattern Recognition Workshops},
  pages={8434--8444},
  year={2026}
}

@inproceedings{mun2020lgi, title={Local-Global Video-Text Interactions for Temporal Grounding}, author={Mun, Jonghwan and Cho, Minsu and Han, Bohyung}, booktitle={CVPR}, year={2020}}

@inproceedings{zeng2020drn, title={Dense Regression Network for Video Grounding}, author={Zeng, Runhao and Xu, Haoming and Huang, Wenbing and Chen, Peihao and Tan, Mingkui and Gan, Chuang}, booktitle={CVPR}, year={2020}}

@inproceedings{soldan2021vlg, title={VLG-Net: Video-Language Graph Matching Network for Video Grounding}, author={Soldan, Mattia and Xu, Mengmeng and Qu, Sisi and Tegner, Jesper and Ghanem, Bernard}, booktitle={Proceedings of the IEEE/CVF International Conference on Computer Vision Workshops}, pages={3224--3234}, year={2021}, url={https://openaccess.thecvf.com/content/ICCV2021W/CVEU/html/Soldan_VLG-Net_Video-Language_Graph_Matching_Network_for_Video_Grounding_ICCVW_2021_paper.html}}

@inproceedings{lei2020xml, title={TVR: A Large-Scale Dataset for Video-Subtitle Moment Retrieval}, author={Lei, Jie and Yu, Licheng and Berg, Tamara L. and Bansal, Mohit}, booktitle={ECCV}, year={2020}}

@inproceedings{radford2021clip, title={Learning Transferable Visual Models From Natural Language Supervision}, author={Radford, Alec and Kim, Jong Wook and Hallacy, Chris and Ramesh, Aditya and Goh, Gabriel and Agarwal, Sandhini and Sastry, Girish and Askell, Amanda and Mishkin, Pamela and Clark, Jack and Krueger, Gretchen and Sutskever, Ilya}, booktitle={ICML}, year={2021}}

@article{mokady2021clipcap, title={ClipCap: CLIP Prefix for Image Captioning}, author={Mokady, Ron and Hertz, Amir and Bermano, Amit H.}, journal={arXiv preprint arXiv:2111.09734}, year={2021}}

@inproceedings{gu2022s4, title={Efficiently Modeling Long Sequences with Structured State Spaces}, author={Gu, Albert and Goel, Karan and Re, Christopher}, booktitle={ICLR}, year={2022}}

@article{ganin2016dann, title={Domain-Adversarial Training of Neural Networks}, author={Ganin, Yaroslav and Ustinova, Evgeniya and Ajakan, Hana and Germain, Pascal and Larochelle, Hugo and Laviolette, Francois and Marchand, Mario and Lempitsky, Victor}, journal={Journal of Machine Learning Research}, volume={17}, number={59}, pages={1--35}, year={2016}}

@inproceedings{lester2021prompt,
  title={The Power of Scale for Parameter-Efficient Prompt Tuning},
  author={Lester, Brian and Al-Rfou, Rami and Constant, Noah},
  booktitle={Proceedings of EMNLP}, year={2021}, pages={3045--3059},
  doi={10.18653/v1/2021.emnlp-main.243},
  url={https://aclanthology.org/2021.emnlp-main.243/}}

@article{zhang2021debias,
  title={Towards Debiasing Temporal Sentence Grounding in Video},
  author={Zhang, Hao and Sun, Aixin and Jing, Wei and Zhou, Joey Tianyi},
  journal={arXiv preprint arXiv:2111.04321}, year={2021},
  url={https://arxiv.org/abs/2111.04321}}

@inproceedings{liu2022e2etad,
 title={An Empirical Study of End-to-End Temporal Action Detection},
 author={Liu, Xiaolong and Bai, Song and Bai, Xiang},
 booktitle={CVPR}, year={2022}, pages={20010--20019},
 url={https://arxiv.org/abs/2204.02932}}

@inproceedings{zhang2023tvp,
 title={Text-Visual Prompting for Efficient 2D Temporal Video Grounding},
 author={Zhang, Yimeng and Chen, Xin and Jia, Jinghan and Liu, Sijia and Ding, Ke},
 booktitle={CVPR}, year={2023}, pages={14794--14804},
 url={https://arxiv.org/abs/2303.04995}}

@inproceedings{ren2024timechat,
 title={{TimeChat}: A Time-sensitive Multimodal Large Language Model for Long Video Understanding},
 author={Ren, Shuhuai and Yao, Linli and Li, Shicheng and Sun, Xu and Hou, Lu},
 booktitle={CVPR}, year={2024}, pages={14313--14323},
 url={https://arxiv.org/abs/2312.02051}}
\end{document}